\documentclass[conference]{IEEEtran}

\PassOptionsToPackage{hyphens}{url}

\usepackage[
  hidelinks,
  pdfusetitle,  
]{hyperref}

\usepackage{cite}
\usepackage{amssymb}
\usepackage{booktabs}
\usepackage{fancyhdr}
\usepackage{flushend}
\usepackage{graphicx}
\usepackage{mathtools}
\usepackage{silence}                    
\usepackage{textcomp}
\usepackage{threeparttable}
\usepackage{xcolor}

\makeatletter
\let\MYcaption\@makecaption
\makeatother
\usepackage[font=footnotesize]{subcaption}
\makeatletter
\let\@makecaption\MYcaption
\makeatother

\usepackage{csquotes}
\MakeOuterQuote{"}

\usepackage[
  capitalize,
  nameinlink,
]{cleveref}

\title{Scalable Human-Machine Point Cloud Compression}

\author{%
  \texorpdfstring{%
    \IEEEauthorblockN{Mateen Ulhaq}
    \IEEEauthorblockA{%
      \textit{School of Engineering Science} \\
      \textit{Simon Fraser University} \\
      Burnaby, BC, Canada \\
      mulhaq@sfu.ca%
    }
    \and
    \IEEEauthorblockN{Ivan V. Baji\'c}
    \IEEEauthorblockA{%
      \textit{School of Engineering Science} \\
      \textit{Simon Fraser University} \\
      Burnaby, BC, Canada \\
      ibajic@ensc.sfu.ca%
    }%
  }{%
    Mateen Ulhaq \and Ivan V. Baji\'c%
  }%
}

\newif\ifarxiv
\arxivtrue

\ifarxiv
  \fancypagestyle{firstpage}{
    \setlength{\headheight}{3.5\normalbaselineskip}
    \setlength{\headsep}{1.5\normalbaselineskip}

    \fancyhead[C]{%
      \footnotesize%
      \textit{Proc. PCS 2024}, Taichung, Taiwan, Jun. 2024 \\
      \vspace{0.8\normalbaselineskip}%
      \scriptsize%
      \copyright{} 2024 IEEE. Personal use of this material is permitted. Permission from IEEE must be obtained for all other uses, in any current or future media, including reprinting/republishing this material for advertising or promotional purposes, creating new collective works, for resale or redistribution to servers or lists, or reuse of any copyrighted component of this work in other works.%
    }
    \fancyfoot{}
  }
\fi

\newcommand{\boldvar}[1]{{\boldsymbol{#1}}}

\newenvironment{subsubfigure}[2][]{%
  \begin{subfigure}[#1]{#2}%
    \stepcounter{subsubfigure}%
}{%
    \addtocounter{subfigure}{-1}%
  \end{subfigure}%
}
\newcounter{subsubfigure}

\newlength{\tablesepskip}

\begin{document}

\maketitle

\ifarxiv
  \thispagestyle{firstpage}
\fi

\begin{abstract}
  Due to the limited computational capabilities of edge devices, deep learning inference can be quite expensive.
  One remedy is to compress and transmit point cloud data over the network for server-side processing.
  Unfortunately, this approach can be sensitive to network factors, including available bitrate.
  Luckily, the bitrate requirements can be reduced without sacrificing inference accuracy by using a machine task-specialized codec.
  In this paper, we present a scalable codec for point-cloud data that is specialized for the machine task of classification, while also providing a mechanism for human viewing.
  In the proposed scalable codec, the "base" bitstream supports the machine task, and an "enhancement" bitstream may be used for better input reconstruction performance for human viewing.
  We base our architecture on PointNet++, and test its efficacy on the ModelNet40 dataset.
  We show significant improvements over prior non-specialized codecs.
\end{abstract}

%


\section{Introduction}
\label{sec:introduction}

Point clouds representing 3D visual data are increasingly being used in many applications, including augmented reality, robotics, and autonomous driving.
Advances in deep learning have led to the development of deep models for performing machine vision tasks on point cloud data, including classification, object detection, and segmentation.
However, most deep learning models are computationally expensive.
This poses a challenge for computationally-limited edge devices that want to perform machine vision tasks, and yet are limited in size, energy consumption, cost, and other factors.

One option for performing a machine task on the edge device is to limit the complexity of the model.
Unfortunately, this usually comes at the cost of model accuracy.
Another option is to transmit the input data to a server for machine analysis.
In this approach, the edge device compresses the input data prior to transmission, often using a codec designed to reconstruct the input for human viewing.
However, such \emph{input reconstruction} codecs are not optimized for machine analysis.
Thus, they often spend large amounts of bits on encoding information that is not relevant to the machine task.
This results in a worse rate-accuracy trade-off than is possible with a more specialized codec.
In situations where a low rate is desired --- for instance, in areas of poor network connectivity, or congested networks --- using a non-specialized codec may result in excessively high machine task latency~\cite{shlezinger2022IOTM}.

In order to improve the rate-accuracy trade-off, we may instead use a codec that is specialized for the machine task.
Such codecs often perform part of the machine task on the edge device itself.
In this hybrid approach, the edge device simultaneously compresses the input and performs part of the machine task.
This allows the model to discard unnecessary information, thus reducing the rate, resulting in a system that is more robust to changing network conditions, and may reduce system latency over a certain range of available bitrates~\cite{bajic2021icassp}.

In~\cite{ulhaq2023pointcloud}, a novel codec for point cloud classification was proposed.
This learned codec, based on PointNet~\cite{qi2016pointnet}, compresses the input point cloud into a highly compressed representation that is intended solely for machine analysis, in this case classification.
This codec was shown to achieve a significantly better rate-accuracy trade-off in comparison with alternative methods using standard codecs that are not specialized for machine analysis.
This was achieved by removing not only statistical redundancy, but also task-irrelevant information, during the compression process.

While the codec in~\cite{ulhaq2023pointcloud} achieves a good rate-accuracy trade-off for point cloud classification, it is not suitable for other purposes.
Most applications involving automated machine-based analysis are expected to run the machine task continuously, but may occasionally require human verification or review.
Hence, it is important to develop codecs that are able to support both tasks --- machine vision and human viewing --- efficiently.
In this paper, we present such a scalable codec, the first in the point cloud literature, which supports point cloud classification, while also providing a mechanism for human viewing.
Our code is available online.%
\footnote{%
  \hfill%
  \url{https://github.com/multimedialabsfu/learned-point-cloud-compression-for-classification}%
}%


\section{Related work}
\label{sec:related-work}

Point cloud classification is among the most researched point cloud analysis tasks.
Related classification models accept different input formats, including
point lists (e.g., PointNet~\cite{qi2016pointnet} and PointNet++~\cite{qi2017pointnetplusplus}),
3D voxel grids (e.g., VoxNet~\cite{maturana2015voxnet}), and
Octrees (e.g., OctNet~\cite{riegler2016octnet}).
Since our work builds on PointNet and PointNet++, we assume the input format is a point list.

Conventional handcrafted point cloud codecs include Draco~\cite{google2017draco} and G\nobreakdash-PCC~\cite{mpeg2019gpccv2} (implemented as TMC13~\cite{mpeg2021tmc13}).
More recently, the research focus has shifted towards learned codecs, following the seminal work of Ball{\'e} \emph{et al.}~\cite{balle2018variational} who proposed a variational autoencoder (VAE) based architecture for image compression.
In their architecture, the input $\boldvar{x}$ is first transformed into a latent representation $\boldvar{y}$, which is then quantized and entropy-coded using a learned entropy model.
Such an architecture can be trained end-to-end using the loss 
\begin{equation}
    \mathcal{L} = R + \lambda \cdot D(\boldvar{x}, \boldvar{\hat{x}}),
    \label{eq:rd-loss}
\end{equation}
where $D(\boldvar{x}, \boldvar{\hat{x}})$ is the distortion measure between the input $\boldvar{x}$ and decoded $\boldvar{\hat{x}}$, and $R$ is the estimate of the entropy of $\boldvar{\hat{y}}$.
This mechanism has been successful in various fields of learned compression, including point cloud compression~%
\cite{%
  quach2019icip,
  yan2019deep,
  he2022density,
  fu2022octattention,
  you2022ipdae
}.


Specialized compression for machine tasks --- often referred to as \emph{coding for machines} --- has been explored for images~\cite{le2021icm} and video~\cite{duan2020vcm} and, recently, for point clouds~\cite{ulhaq2023pointcloud}.
Moreover, scalable multi-task coding approaches~\cite{hu2020towardscfhmvscalable,choi2022sichm} have shown that one can perform a machine vision task at a fairly low bitrate, while enabling other tasks, such as input reconstruction for human viewing, with an additional enhancement bitstream.

While quality-scalable point cloud coding has been studied before~\cite{PC_scalable_ICIP2020}, this paper presents the first scalable multi-task point cloud codec: the base bitstream supports point cloud classification, while the enhancement bitstream allows point cloud reconstruction.
Unlike our earlier work~\cite{ulhaq2023pointcloud}, which presented a classification-optimized codec based on PointNet~\cite{qi2016pointnet}, our scalable codec is based on PointNet++~\cite{qi2017pointnetplusplus}, which is a hierarchical extension of PointNet.

\section{Proposed codec}
\label{sec:proposed-codec}

\begin{figure}[tbp]
  \centering
  \captionsetup[subfigure]{aboveskip=0pt,belowskip=\baselineskip}
  \begin{subfigure}[b]{\linewidth}
    \centering
    \includegraphics[width=\linewidth]{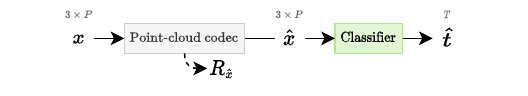}
    \caption{Input compression.}
    \label{fig:arch-comparison/input-compression}
  \end{subfigure}%
  \par%
  \begin{subfigure}[b]{\linewidth}
    \centering
    \includegraphics[width=\linewidth]{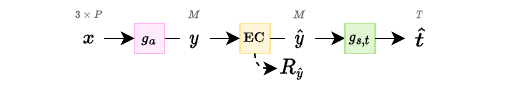}
    \caption{Machine task compression, as used in e.g.~\cite{ulhaq2023pointcloud}.}
    \label{fig:arch-comparison/machine-task}
  \end{subfigure}%
  \par%
  \begin{subfigure}[b]{\linewidth}
    \centering
    \includegraphics[width=\linewidth]{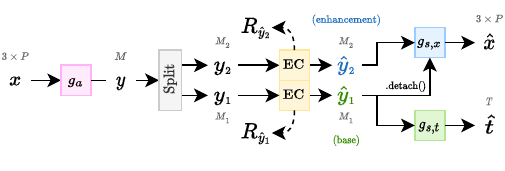}
    \caption{Scalable multi-task compression.}
    \label{fig:arch-comparison/scalable-multi-task}
  \end{subfigure}%
  \vspace{-0.8\baselineskip}%
  \caption{High-level comparison of codec architectures.}
  \label{fig:arch-comparison}
  \vspace{-0.5\baselineskip}
\end{figure}

\begin{figure*}[t]
  \centering
  \includegraphics[width=1.0\linewidth]{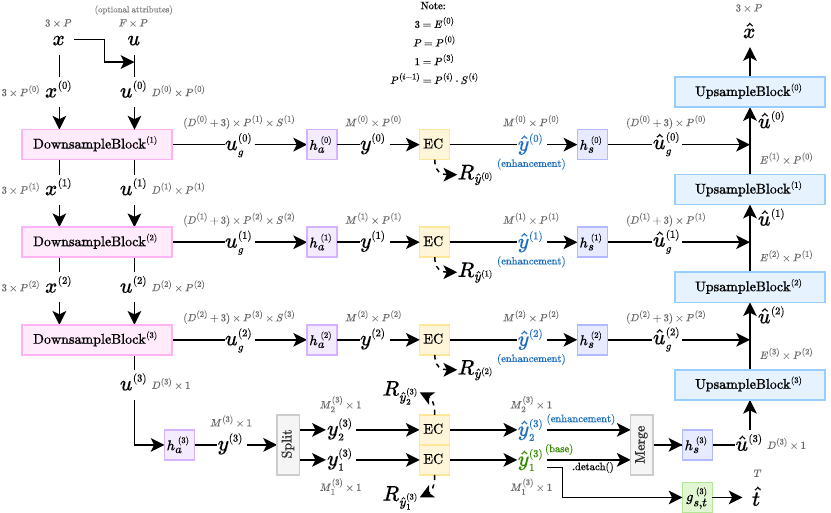}
  \caption{Proposed codec architecture.}
  \label{fig:arch-proposed-full}
\end{figure*}

\begin{figure}[htbp]
  \centering
  \includegraphics[width=1.0\linewidth]{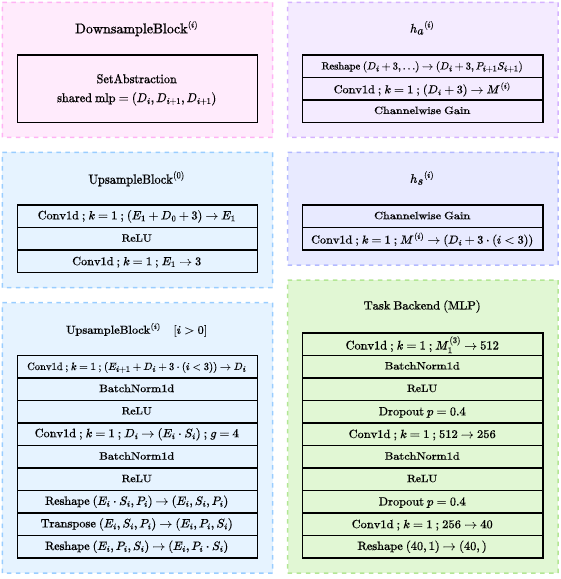}
  \caption{Proposed codec architecture (details).}
  \label{fig:arch-proposed-full-blocks}
\end{figure}

\subsection{Preliminaries}

In \cref{fig:arch-comparison/input-compression}, we show an abstract representation of an \emph{input compression codec}. 
The input point cloud $\boldvar{x}$ is encoded and decoded as $\boldvar{\hat{x}}$ by any desired point cloud codec, including non-learned codecs such as G-PCC~\cite{mpeg2019gpccv2}.
Then, the reconstructed point cloud $\boldvar{\hat{x}}$ is fed into a classification model (e.g., PointNet) in order to obtain the class prediction $\boldvar{\hat{t}}$.
This approach provides a baseline for comparison with our proposed codec.

In \cref{fig:arch-comparison/machine-task}, we show an abstract representation of a \emph{machine task codec}, as was explored by~\cite{ulhaq2023pointcloud} for point cloud classification.
Using the same terminology as in~\cite{balle2018variational}, $g_a$ refers to the \emph{analysis transform}, and $g_s$ refers to the \emph{synthesis transform}.
In this codec, the input point cloud $\boldvar{x}$ is first encoded into a latent representation $\boldvar{y} = g_a(\boldvar{x})$.
This is then quantized as $\boldvar{\hat{y}} = Q(\boldvar{y})$, and then losslessly compressed using a learned entropy model.
For instance, in~\cite{ulhaq2023pointcloud}, a fully-factorized entropy model was used.
The reconstructed latent representation $\boldvar{\hat{y}}$ may then be used to predict the classes $\boldvar{\hat{t}} = g_{s,t}(\boldvar{\hat{y}})$.

\subsection{Scalable human-machine compression codec}
\label{sec:proposed-codec/scalable-multi-task-codec}

In \cref{fig:arch-comparison/scalable-multi-task}, we show a high-level representation of a \emph{scalable multi-task codec}.
Following the principle of \emph{latent space scalability}~\cite{choi2022sichm}, the scalable multi-task codec splits the latent space into two parts, $[\boldvar{\hat{y}}_1, \boldvar{\hat{y}}_2] = \operatorname{split}(\boldvar{\hat{y}})$, along the channel dimension.
The first part $\boldvar{\hat{y}}_1$ is called the "base" layer and the second part $\boldvar{\hat{y}}_2$ is called the "enhancement" layer.
The base layer is used for the machine task (i.e., classification) to predict the class $\boldvar{\hat{t}} = g_{s,t}(\boldvar{\hat{y}}_1)$.
Both base and enhancement layers are used for input reconstruction:
$\boldvar{\hat{x}} = g_{s,x}(\operatorname{concat} [\operatorname{detach}(\boldvar{\hat{y}}_1), \boldvar{\hat{y}}_2])$.
This approach allows for scalability: the enhancement bitstream only needs to be computed and transmitted when human viewing is desired.
Note that the $\operatorname{detach}$ operation is used to disable gradient propagation of $D(\boldvar{x}, \boldvar{\hat{x}})$ backwards through $\boldvar{\hat{y}}_1$.
This improves the specialization of $\boldvar{\hat{y}}_1$ towards the machine task; otherwise, $\boldvar{\hat{y}}_1$ may end up with some enhancement-layer information~\cite{foroutan2023icip}.
Two separate bitstreams (base and enhancement) are produced, whose total rate is
$R_{\boldvar{\hat{y}}} = R_{\boldvar{\hat{y}}_1} + R_{\boldvar{\hat{y}}_2}$.

\subsection{Proposed architecture}

Our complete proposed codec architecture, based on PointNet++~\cite{qi2017pointnetplusplus}, is shown in~\cref{fig:arch-proposed-full},
and the details of each block are shown in~\cref{fig:arch-proposed-full-blocks}.
Our codec is provided in "full" and "lite" configurations, as detailed in~\cref{tbl:layers}.
The input point cloud of $P$ points is represented as a matrix $\boldvar{x} \in \mathbb{R}^{3 \times P}$ of $(x, y, z)$ coordinates.
$\boldvar{u}$ is an additional set of attribute features (e.g., normals, color, etc.) that optionally may also be compressed.
At the beginning of the encoder, $\boldvar{x}$ and $\boldvar{u}$ are concatenated along the channel dimension so that $\boldvar{u}^{(0)} = \operatorname{concat} [\boldvar{x}, \boldvar{u}]$.
This ensures that the same encoding capabilities are available for both.
We feed $\boldvar{x}^{(0)} = \boldvar{x}$ and $\boldvar{u}^{(0)}$ into a sequence of downsampling blocks.
Each $i$-th downsampling block takes in $\boldvar{x}^{(i-1)}$ and $\boldvar{u}^{(i-1)}$,
and outputs and a smaller set of centroids $\boldvar{x}^{(i)}$ and features $\boldvar{u}^{(i)}$,
along with features $\boldvar{u_g}^{(i-1)}$ that are grouped by the centroids.
The final $\boldvar{u}^{(3)}$ is compressed in a multi-task scalable manner, from which $\boldvar{\hat{u}}^{(3)}$ is derived.
Each $\boldvar{u_g}^{(i-1)}$ is compressed using a standard transform-encoder-decoder-transform compression pipeline to obtain $\boldvar{\hat{u}_g}^{(i-1)}$.
A sequence of upsampling blocks is then applied, where each $i$-th upsampling block takes in $\boldvar{\hat{u}}^{(i)}$.
For $i > 0$, the output of the $i$-th upsampling block is concatenated with $\boldvar{\hat{u}_g}^{(i-1)}$ to give $\boldvar{\hat{u}}^{(i-1)}$.
For $i = 0$, the output is $\boldvar{\hat{x}}$.



\subsubsection{Downsampling}

Each downsampling block is a PointNet++ "set abstraction" layer (see~\cite{qi2017pointnetplusplus} for more details), with minor modifications.
We used single-scale grouping (SSG) of points for our proposed codec, though it may likely be improved with multi-scale grouping (MSG) or multi-resolution grouping (MRG) as described in~\cite{qi2017pointnetplusplus}.
The "set abstraction" layer takes as input a (subsampled) point cloud $\boldvar{x}^{(i-1)}$ of shape $3 \times P^{(i-1)}$ and features $\boldvar{u}^{(i-1)}$ of shape $D^{(i-1)} \times P^{(i-1)}$ containing information about each corresponding point.
Using farthest point sampling (FPS), a set of $P^{(i)}$ centroids $\boldvar{x}^{(i)}$ of shape $3 \times P^{(i)}$ is selected from $\boldvar{x}^{(i-1)}$.
Then, the points in $\boldvar{x}^{(i-1)}$ are grouped into $P^{(i)}$ groups of $S^{(i)}$ points each.
Then for each centroid, a ball query\footnote{%
  Note that due to the ball query, the same point may be assigned to multiple centroids.
  Also, centroids are not always present within their own group of points.
  Nonetheless, the ball query has the benefit of scale-invariant grouping.%
}~\cite{qi2017pointnetplusplus}
is performed to find the first $S^{(i)}$ points that are within a radius of $R^{(i)}$ from the given centroid.
The relative positions of each group of points is then computed with respect to its associated centroid, resulting in the residuals
$r^{(i)} = \operatorname{ball\_query}(\boldvar{x}^{(i-1)}, \boldvar{x}^{(i)}, R^{(i)}) - \operatorname{repeat}(\boldvar{x}^{(i)}, S^{(i)})$
of shape $3 \times P^{(i)} \times S^{(i)}$.
Each point in $\boldvar{r}^{(i)}$ is concatenated with its respective feature vector in $\boldvar{u}^{(i-1)}$, resulting in a grouped feature tensor $\boldvar{u_g}^{(i-1)}$ of shape $(D^{(i-1)} + 3) \times P^{(i)} \times S^{(i)}$.
Then, $\boldvar{u_g}^{(i-1)}$ is fed into a miniature group-wise "PointNet encoder" block consisting of a shared multi-layer perceptron (MLP) with a max pooling operation at the end.
That is, the same PointNet encoder is applied to each group of features independently and identically.
This results in $\boldvar{u}^{(i)}$ of shape $D^{(i)} \times P^{(i)}$.
%
Three downsampling blocks are used in our proposed codec.
The last downsampling block groups all the remaining points into a single group, i.e., $P^{(3)} = 1$.


\subsubsection{Feature compression}

At each level $i < 3$, a latent representation $\boldvar{y}^{(i)} = {h_a}^{(i)}(\boldvar{u_g}^{(i)})$ of shape $M^{(i)} \times P^{(i)}$ is computed, then quantized as $\boldvar{\hat{y}}^{(i)} = Q(\boldvar{y}^{(i)})$, and then entropy-coded using a learned entropy model.
The resulting bitstream is transmitted and then decoded as $\boldvar{\hat{y}}^{(i)}$.
From this, a set of grouped features $\boldvar{\hat{u}_g}^{(i)} = h_s^{\, (i)}(\boldvar{\hat{y}}^{(i)})$ is computed for usage during upsampling.
(Note that $\boldvar{\hat{u}_g}^{(i)}$ and $\boldvar{u_g}^{(i)}$ may inhabit entirely different feature spaces, and are unrelated.)

The final feature vector $\boldvar{u}^{(3)}$ is fed into the scalable multi-task compression pipeline described in~\cref{sec:proposed-codec/scalable-multi-task-codec}.
The latent representation $\boldvar{y}^{(3)} = {h_a}^{(3)}(\boldvar{u}^{(3)})$ is computed, then quantized as $\boldvar{\hat{y}}^{(3)} = Q(\boldvar{y}^{(3)})$, and then entropy-coded to generate a base bitstream for $\boldvar{\hat{y}}_1^{(3)}$ and an enhancement bitstream for $\boldvar{\hat{y}}_2^{(3)}$.
The base bitstream is decoded as $\boldvar{\hat{y}}_1^{(3)}$, and then fed into a decoder-side task backend to obtain the class prediction $\boldvar{\hat{t}} = g_{s,t}^{(3)}(\boldvar{\hat{y}}_1^{(3)})$.
The enhancement bitstream is decoded as $\boldvar{\hat{y}}_2^{(3)}$, and then its concatenation with $\boldvar{\hat{y}}_1^{(3)}$ (detached) is fed into a decoder-side transform to obtain $\boldvar{\hat{u}_g}^{(3)} = h_s^{(3)}(\operatorname{concat} [\operatorname{detach}(\boldvar{\hat{y}}_1^{(3)}), \boldvar{\hat{y}}_2^{(3)}])$.

Note that we split only $\boldvar{y}^{(3)}$ into $\boldvar{y}^{(3)}_1$ and $\boldvar{y}^{(3)}_2$.
This is because $\boldvar{y}^{(3)}_1$ suffices for the classification task, while $\boldvar{y}^{(3)}_2$ and $\boldvar{y}^{(j)}$ for $j<3$ are only needed for input reconstruction.
For brevity, in the experiments we only test input reconstruction with all enhancement layers, but the proposed architecture also allows for quality- or density-scalable input reconstruction.



\subsubsection{Upsampling}

An upsampling block takes in $E^{(i+1)} + D^{(i)} + 3 \cdot (1 - \delta_{i,3})$ channels and outputs $E^{(i)}$ channels, where $\delta_{i,j}$ is the Kronecker delta ($\delta_{i,j}=1$ if $i=j$, $0$ otherwise).
Each upsampling block consists of a simple two-layer point-wise MLP.
For upsampling blocks at levels deeper than $i > 0$, batch normalizations are also present.
Furthermore, each such block is followed by a reshape-transpose-reshape combination that converts the resulting shape
$(E^{(i)} \cdot S^{(i)}) \times P^{(i)}$
into the output shape
$E^{(i)} \times (S^{(i)} \cdot P^{(i)})$.
To ensure compatibility of shapes among various components during reconstruction, we enforce the constraint $P^{(i-1)} = P^{(i)} \cdot S^{(i)}$.



\begin{table}[t]
  \centering
  \caption{Architecture hyperparameter configurations}
  \label{tbl:layers}
  \scriptsize
  \setlength{\tablesepskip}{-0.9\normalbaselineskip}
  \begin{tabular}[]{cccccccc}
    \toprule
    \\[\tablesepskip]
    Codec & $i$ & $P^{(i)}$ & $S^{(i)}$ & $R^{(i)}$ & $D^{(i)}$ & $E^{(i)}$ & $M^{(i)}$ \\
    \\[\tablesepskip]
    \midrule
    \\[\tablesepskip]
    full  & 0 & 1024 &    &     & 3   & 3  & 0 \\
          & 1 & 256  & 4  & 0.2 & 128 & 64 & 0 \\
          & 2 & 64   & 4  & 0.4 & 192 & 32 & 64 \\
          & 3 & 1    & 64 &     & 256 & 16 & (48, 16) \\
    \\[\tablesepskip]
    \midrule
    \\[\tablesepskip]
    lite  & 0 & 1024 &    &     & 3  & 3  & 0 \\
          & 1 & 256  & 4  & 0.2 & 32 & 32 & 0 \\
          & 2 & 64   & 4  & 0.4 & 48 & 16 & 16 \\
          & 3 & 1    & 64 &     & 64 & 8  & (48, 16) \\
    \\[\tablesepskip]
    \bottomrule
  \end{tabular}
\end{table}

\section{Experiments}

We trained our models on the ModelNet40~\cite{wu20143d} dataset,
sampling $P = 1024$ points per object, and reconstructed the same number of points.
Our implementation uses the PyTorch, CompressAI~\cite{begaint2020compressai}, and CompressAI Trainer~\cite{ulhaq2022compressaitrainer} libraries.


\begin{figure*}[htbp]
  \centering
  \begin{subfigure}{0.30\linewidth}
    \centering
    \includegraphics[width=\linewidth]{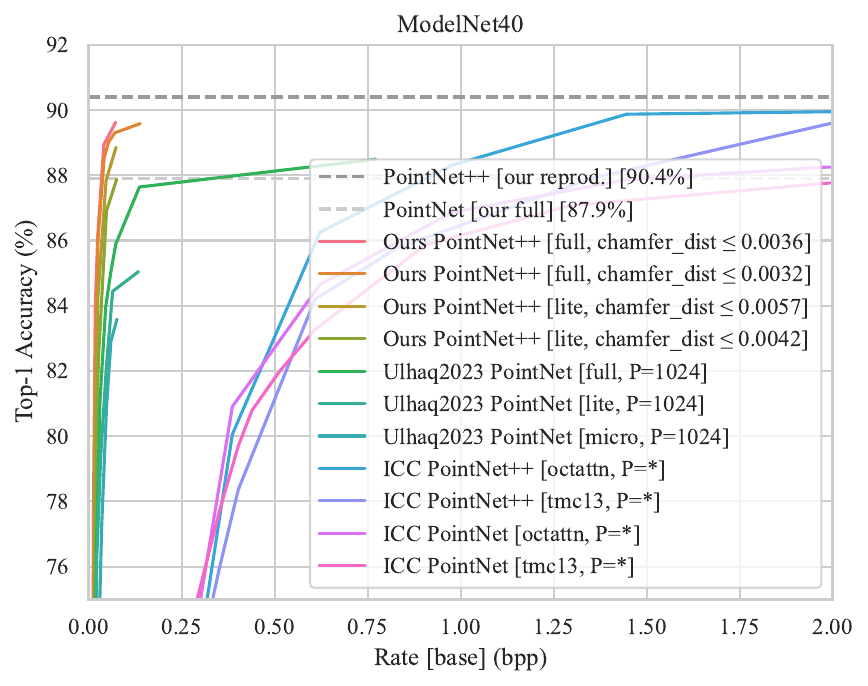}
    \caption{RA for base task.}
    \label{fig:ra-rd-curves/basetask-unzoomed}
  \end{subfigure}%
  \begin{subfigure}{0.30\linewidth}
    \centering
    \includegraphics[width=\linewidth]{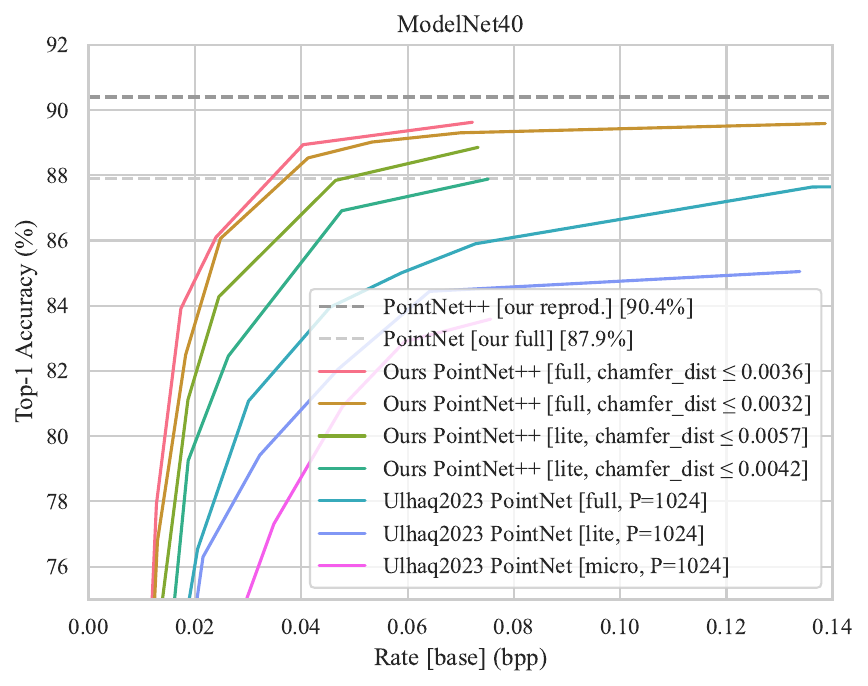}
    \caption{RA for base task (zoomed in).}
    \label{fig:ra-rd-curves/basetask-zoomed}
  \end{subfigure}%
  \begin{subfigure}{0.30\linewidth}
    \centering
    \includegraphics[width=\linewidth]{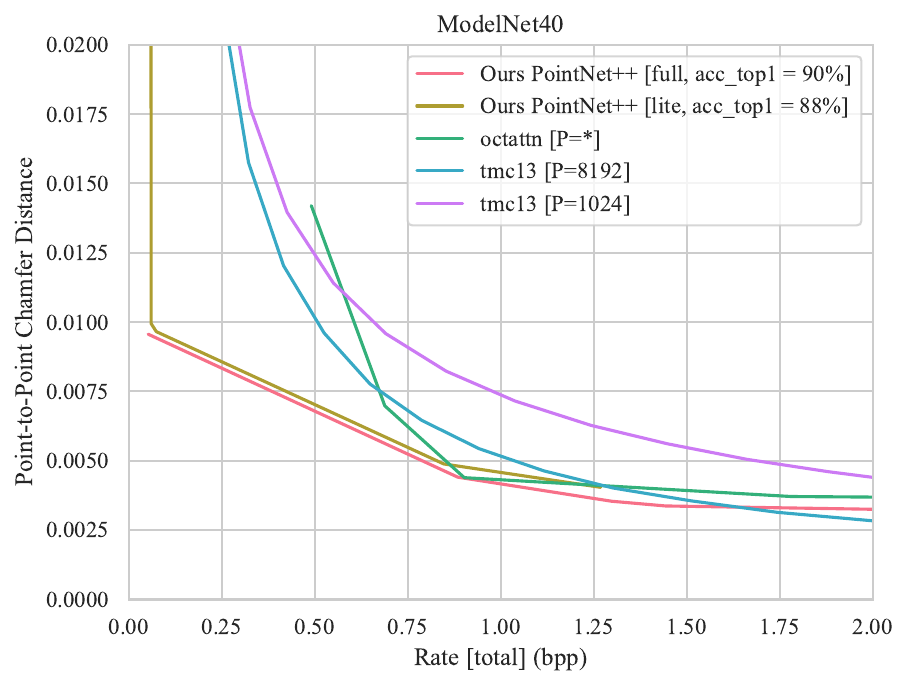}
    \caption{RD for reconstruction task.}
    \label{fig:ra-rd-curves/rec-chamfer}
  \end{subfigure}%
  \caption{%
    Rate-accuracy (RA) and rate-distortion (RD) curves on the ModelNet40 dataset,
    with rate units of bits per point (bpp) scaled for 1024 points.
  }
  \label{fig:ra-rd-curves}
\end{figure*}

The loss function used for training was
\begin{equation}
  \mathcal{L}
  =
  \sum_{i=0}^{3} R_{\boldvar{\hat{y}}^{(i)}}
  + \lambda_{x} D(\boldvar{x}, \boldvar{\hat{x}})
  + \lambda_{t} D(\boldvar{t}, \boldvar{\hat{t}}).
\end{equation}
The rate estimate
$R_{\boldvar{\hat{y}}^{(i)}}$ 
for each level is the negative log likelihood outputted by the respective entropy models.
The distortion $D(\boldvar{t}, \boldvar{\hat{t}})$ is the cross-entropy between the one-hot encoded labels $\boldvar{t}$ and the softmax of the model's prediction $\boldvar{\hat{t}}$.
The distortion $D(\boldvar{x}, \boldvar{\hat{x}})$ is the Chamfer distance between the input and reconstructed point clouds.
We trained different models to operate at different base/enhancement rate points by varying the hyperparameters
$\lambda_{t} \in \{ 2^{-7}, 2^{-6}, \ldots, 2^{0} \}$
and
$\lambda_{x} \in [1, 8000]$.
These values are reported for rates in units of bits per point (bpp), i.e., the rates are divided by $P = 1024$.
To improve stability during training, we disabled the first two levels by setting $M^{(0)} = 0$ and $M^{(1)} = 0$.
We also set $M^{(2)} = 64$ and $M^{(3)} = 64$, and used a partitioning ratio of $0.75$ for the last layer, i.e., $M^{(3)}_1 = 48$ (base) and $M^{(3)}_2 = 16$ (enhancement).

The input compression codecs were evaluated using the same methodology for varying $P$ and input scaling as in~\cite{ulhaq2023pointcloud}.
Since codecs like TMC13~\cite{mpeg2021tmc13} and OctAttention~\cite{fu2022octattention} are tailored to point clouds with a large number of points, the coding overhead they produce may be significant when applied to smaller point clouds such as those used in classification.
To improve the fairness when measuring the rate, we ignored such overhead within the bitstream when possible.

\section{Results}


\cref{fig:ra-rd-curves/basetask-zoomed,fig:ra-rd-curves/basetask-unzoomed} show the rate-accuracy curves for the base task of point-cloud classification.
Only the rate of the base bitstream is measured for the base task.
Scalable codec curves are grouped by their worst-case reconstruction ability measured in terms of Chamfer distance~\cite{blum1967chamfer}.
Our proposed codec achieves better classification performance than all existing codecs,
including~\cite{ulhaq2023pointcloud} (denoted "Ulhaq2023"),
which in turn is better than input compression codecs ("ICC") that use a point-cloud codec (TMC13~\cite{mpeg2021tmc13} and OctAttention~\cite{fu2022octattention}) followed by a PointNet/PointNet++ classifier.
Only the best ICC curves are shown.
Note that "$P{=}{}^*$" indicates the best results attained when varying $P \in \{8, 16, \ldots, 1024\}$.

There is a large gap in base task performance between task-specialized codecs (the proposed one and~\cite{ulhaq2023pointcloud}) and non-specialized codecs because, in addition to statistical redundancy, the former codecs remove a significant amount of task-irrelevant information.
Meanwhile, the proposed codec outperforms~\cite{ulhaq2023pointcloud} due to its reliance on the more powerful PointNet++, rather than PointNet.

\cref{fig:ra-rd-curves/rec-chamfer} show the rate-distortion curves for the reconstruction task in terms of the rate versus the Chamfer distance. 
For the proposed codec, the rate includes both the base and enhancement bitstreams.
The figures show that our proposed codec is also capable of achieving better reconstruction quality than conventional codecs at low rates, below 1.6 bpp. 
As the rate increases, the conventional codecs eventually catch up in terms of reconstruction quality.

\section{Conclusion}

In this paper, we proposed a scalable multi-task compression codec for point clouds.
Our proposed codec produces a base bitstream that supports point cloud classification and an enhancement bitstream that, together with the base bitstream, supports reconstruction of the point cloud for human viewing. 
The proposed codec showed a significant improvement in rate-accuracy performance for the base task over existing codecs, while also achieving competitive performance against conventional codecs on input reconstruction at reasonably low rates.
In the future, we hope that this work will inspire further research into scalable multi-task compression codecs for point clouds on more complex real-world datasets, and for other tasks such as semantic segmentation.


%


\phantomsection

%
\bibliographystyle{IEEEtran}
\bibliography{references}


\end{document}